\documentclass{article} % For LaTeX2e

\usepackage{arxiv_preprint,times}
\usepackage{hyperref}
\usepackage{url}
\usepackage{booktabs}
\usepackage{graphicx}
\usepackage{amsmath}
\usepackage{amssymb}
\usepackage{array}
\usepackage{makecell}
\usepackage{pifont}
\usepackage{wrapfig}
\title{Chinese Competitive Debate: A Dataset and Benchmark with Professional Adjudication}

\author{Zongrui Yang, Haoyuan Li, Zhongsheng Wang, Zhirui Zeng,\\
Pengqian Han, Yi Zhou, Yuting Wang, Jiamou Liu}

\begin{document}

\maketitle

\begin{abstract}
We introduce a Chinese competitive debate dataset and benchmark that links transcripts to professional assessments of local performance and overall outcomes. Existing resources provide valuable transcripts and annotations, but access to original stage-level judgments collected during real competitions under a shared rubric remains limited. We organized 182 matches, each independently adjudicated by three qualified judges using a predefined rubric. After filtering incomplete records, the dataset contains 148 unique matches, 2,698 stage-level instances, and 20,542 exchange units. It preserves manually reviewed transcripts, individual stage scores, match votes, best-debater ballots, and original adjudication commentary. Three prediction tasks cover match outcomes, stage scores, and best-debater selections. In initial zero-shot evaluations, the highest winner-prediction accuracy is 66.2\%, the highest stage-score Pearson correlation is 0.250, and the highest best-debater accuracy is 56.8\%. Judge-agreement analyses and structural baselines contextualize these results. The resource supports reproducible evaluation of model agreement with professional adjudication across multiple levels of a debate.

\end{abstract}

\section{Introduction}

Debate adjudication involves assessing contributions in relation to the preceding interaction and the speaker's role. A response may be fluent and relevant to the motion yet leave the opponent's objection unanswered. Studying how such contributions are assessed requires records of both the debate and the judgments associated with it. Complete transcripts preserve what participants said, while stage-level scores record how judges evaluated particular performances. Linking these records provides supervision for developing and evaluating automated adjudication systems.

Prior work has investigated automated adjudication of complete debates and the evaluation of individual speeches, including PanelBench \citep{debatrix2024}, DebateBench \citep{tiwari2025debatebench}, and Debatable Intelligence \citep{debatable2025}. These resources establish important foundations for comparing model predictions with human judgments. The available supervision nevertheless depends on the debate format and the assessment procedure. We focus on Chinese competitive debate, whose formats combine constructive and closing speeches with questioning, cross-examination, head-to-head exchanges, and free debate. These stages impose different obligations on participants: a questioner develops challenges, a respondent addresses them, and a closing speaker synthesizes the comparison between the two sides. Role-specific stage assessments therefore provide information beyond the final match outcome. Appendix~\ref{app:debate-formats} discusses the relevant format differences.

Existing Chinese debate datasets offer different forms of supervision: ORCHID \citep{zhao2023orchid} provides transcripts with stance and summarization annotations; CEDAR \citep{lan2026cedar} includes argument-related annotations, speaker roles, timing, outcomes, and text and audio; Chen et al.~\citep{chen2026debate} and DEFINED \citep{yu2026defined} support debate assessment; and Conch \citep{chen2026conch} analyzes clash structure in three debates. Access remains limited to original stage-level scores collected under a shared rubric and linked to individual judges' decisions, speaker ballots, and commentary in the same competition record. Such supervision is difficult to reconstruct from transcripts and published outcomes alone. Table~\ref{tab:dataset-comparison} compares these resources.

\begin{table*}[t]
\centering
% Local symbol definitions.
\def\cmark{\ensuremath{\checkmark}}
\def\xmark{\ensuremath{\times}}
\def\pmark{\ensuremath{\triangle}}
\def\nmark{--}

\small
\setlength{\tabcolsep}{3pt}
\renewcommand{\arraystretch}{1.15}

\caption{
Comparison of selected debate datasets and adjudication benchmarks.
\cmark: available;
\pmark: partially available;
\xmark: unavailable;
\nmark: not applicable.
}
\label{tab:dataset-comparison}

\resizebox{\textwidth}{!}{%
\begin{tabular}{@{} l cc ccc ccc ccc cc @{}}
\toprule

& \multicolumn{2}{c}{Basic info}
& \multicolumn{3}{c}{Text organization}
& \multicolumn{3}{c}{Supervision}
& \multicolumn{3}{c}{Stage-level scoring}
& \multicolumn{2}{c}{\makecell{Release and\\privacy}} \\

\cmidrule(lr){2-3}
\cmidrule(lr){4-6}
\cmidrule(lr){7-9}
\cmidrule(lr){10-12}
\cmidrule(lr){13-14}

Dataset
& Source
& Retained debates
& Stage
& Clash
& \makecell{Judge \\commentary}
& Outcome
& Stage scores
& \makecell{Best\\Debater}
& \makecell{Competition-\\time adjudica.}
& \makecell{Unified\\rubric}
& Rater
& Public
& Anonymized \\

\midrule

\makecell[l]{ORCHID\\\citep{zhao2023orchid}}
& Re-curated
& 1,218
& \cmark & \xmark & \xmark
& \xmark & \xmark & \xmark
& \nmark & \nmark & \nmark
& \cmark & \cmark \\
\addlinespace[3pt]

\makecell[l]{Chen et al.\\\citep{chen2026debate}}
& Re-curated
& 94
& \cmark & \cmark & \xmark
& \cmark & \xmark & \xmark
& \xmark & \xmark & LLM
& \xmark & \xmark \\
\addlinespace[3pt]

\makecell[l]{DEFINED\\\citep{yu2026defined}}
& Re-curated
& 108
& \cmark & \xmark & \xmark
& \xmark & \pmark & \xmark
& \xmark & \cmark & \makecell{Expert/\\student}
& \xmark & \xmark \\
\addlinespace[3pt]

\makecell[l]{CEDAR\\\citep{lan2026cedar}}
& Re-curated
& 600
& \pmark & \xmark & \pmark
& \cmark & \xmark & \pmark
& \nmark & \nmark & \nmark
& \cmark & \xmark \\
\addlinespace[3pt]

\makecell[l]{Conch\\\citep{chen2026conch}}
& Re-curated
& 3
& \cmark & \cmark & \xmark
& \xmark & \xmark & \xmark
& \nmark & \nmark & \nmark
& \pmark & \pmark \\

\midrule

\textbf{Ours}
& \makecell{Self-\\organized}
& \textbf{148}
& \cmark & \cmark & \cmark
& \cmark & \cmark & \cmark
& \cmark & \cmark & Expert
& \cmark & \cmark \\

\bottomrule
\end{tabular}%
}

\par\smallskip
\begin{minipage}{\textwidth}
\footnotesize
\textit{Notes.}
Competition-time adjudication denotes judgments collected as part of the competition procedure rather than annotations added solely for subsequent dataset construction. A shared rubric denotes an explicit scoring framework applied within the dataset. We collected 182 matches and retained 148 with complete scoring records.
\end{minipage}

\end{table*}

To address this gap, we introduce the \textbf{Chinese Competitive Debate Dataset (CCDD)} and an accompanying adjudication benchmark. Before organizing the competitions, we developed a shared scoring rubric with senior debate experts. We then collected 182 matches, each independently assessed by three judges selected from a pool of 157 qualified adjudicators. The collection procedure preserved stage scores, match votes, best-debater ballots, and post-match commentary. Organizing the competitions enabled us to capture these original adjudication records systematically and align them with the debate transcripts.

\begin{figure}[htbp]
    \centering
    \includegraphics[width=\linewidth]{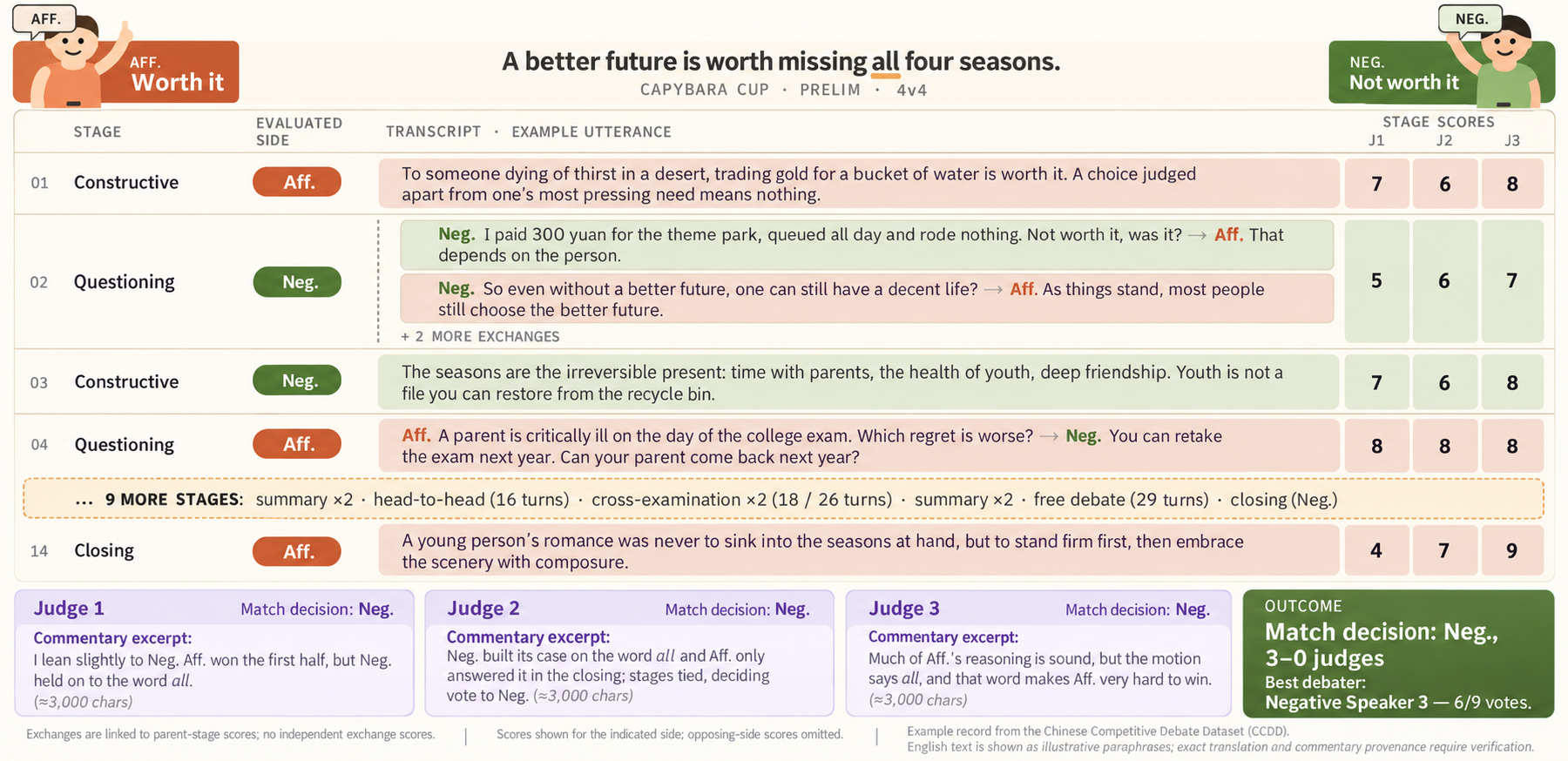}
    \caption{Example record from the CCDD. Selected stages and illustrative English paraphrases are shown alongside three judges' scores for the indicated side. Exchanges are linked to their parent stages and do not have independent quality labels. The lower panels illustrate accompanying judge commentary, the match decision, and the best-debater result.}
    \label{fig:dataset}
\end{figure}

After excluding matches with incomplete scoring records, CCDD contains 148 unique matches, 2,698 stage-level instances, and 20,542 exchange units. Transcripts and stage segmentation were manually reviewed, and scores were aligned with the evaluated stages and sides. Individual judges' assessments are preserved alongside aggregate targets, allowing users to examine both consensus and disagreement. Exchange units retain local interaction text and are linked to their parent stages; they do not have independently assigned human quality scores. Figure~\ref{fig:dataset} illustrates these relationships in an example record.

The CCDD benchmark defines three tasks: \textbf{winner prediction}, \textbf{stage-score prediction}, and \textbf{best-debater prediction}. These tasks evaluate agreement with professional judgments at the match, stage, and speaker levels. Initial zero-shot evaluations achieve a highest winner-prediction accuracy of 66.2\%, a highest stage-score Pearson correlation of 0.250, and a highest best-debater accuracy of 56.8\%. We report judge-agreement statistics and structural references to contextualize these results. Together, the dataset and benchmark provide a foundation for studying automated adjudication under an explicit competition rubric.

Our main contributions are as follows:

\textbf{Original professional adjudication records.} CCDD preserves individual stage scores, match votes, best-debater ballots, and commentary collected under a predefined shared rubric during genuine competitions.\quad \textbf{Aligned and documented debate data.} Manually reviewed transcripts are organized into matches, stages, and exchange units and aligned with adjudication records; original and appeal judgments remain distinct, as do original human annotations and supplementary model-generated materials.\quad \textbf{A multilevel adjudication benchmark.} Three prediction tasks are accompanied by initial zero-shot results, judge-agreement analyses, and structural references for evaluating automated debate adjudication.

\section{Dataset Construction and Benchmark Design}
\label{sec:benchmark-construction}

Our pipeline develops a shared rubric, organizes and records competitions, transcribes, corrects, segments, and aligns records with judge assessments, and constructs match-, stage-, and speaker-level benchmark instances. Exchange units additionally organize the interaction text.

\begin{figure}[htbp]
    \centering
    \includegraphics[width=\linewidth]{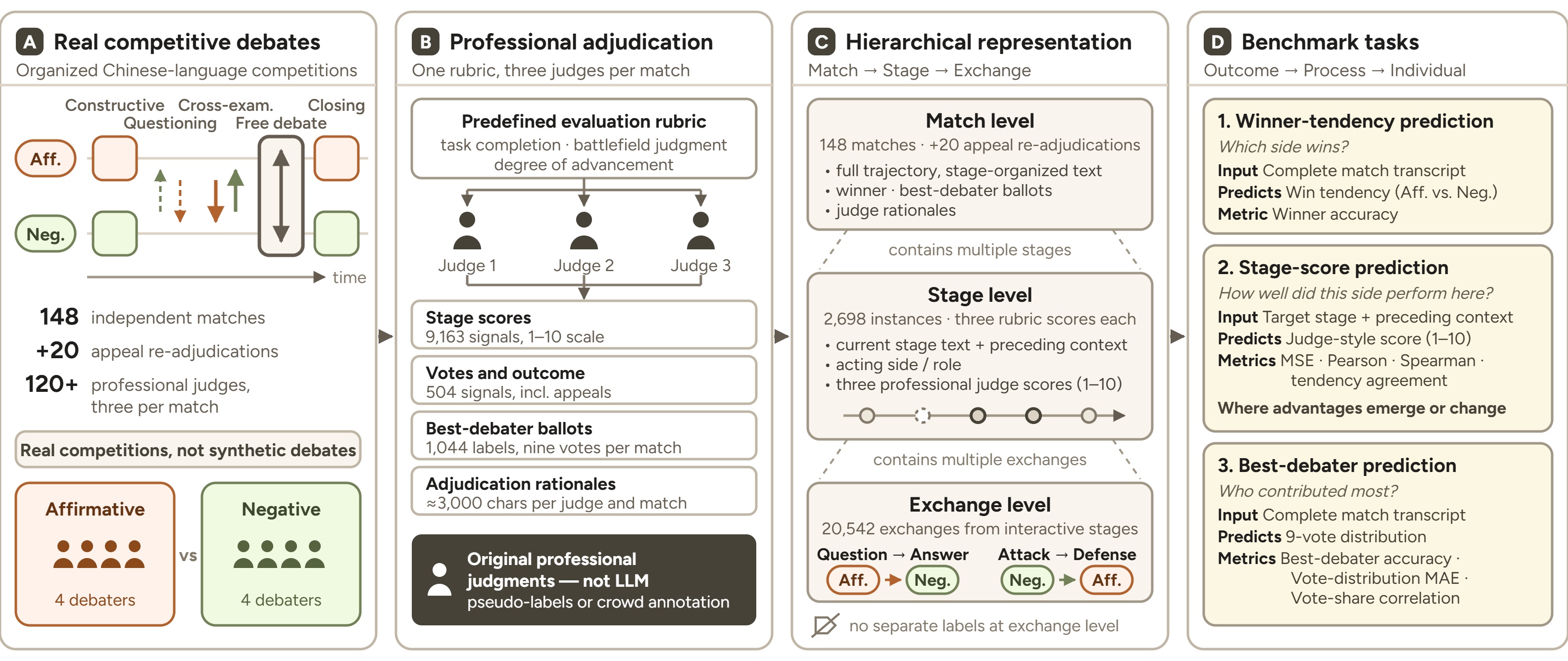}
    \caption{Overview of the Chinese Competitive Debating Dataset and Benchmark.}
    \label{fig:flow}
\end{figure}

\subsection{Debate Format and Evaluation Rubric}
\label{sec:format-and-rubric}

\subsubsection{Debate Format}

Competitive debate in mainland China follows formats represented
by tournaments such as Xin Guo Bian, the Chinese Debate World Cup,
and the Huaxia Cup. Although these tournaments differ in speaker
assignments and timing rules, they share broadly similar
match structures.

We divide the 4v4 format used in our competitions into three
\emph{phases} according to their primary functions.
The constructive phase establishes the argumentative frameworks
through constructive speeches, questioning, and questioning
summaries.
The clash phase develops attacks and defenses on central points
of contention through head-to-head exchanges, cross-examination,
and cross-examination summaries.
The concluding phase includes free debate and closing speeches,
allowing further exchanges and synthesis of the overall comparison.
Each phase comprises multiple individual \emph{stages}, which
serve as basic units of scoring and evaluation.
Stage order, speaker roles, timing rules, and a diagram of the
competition procedure are provided in
Appendix~\ref{app:chinese-debate-format}.

These stages impose different interaction constraints.
Speech stages involve uninterrupted delivery by one side.
Questioning and cross-examination follow an asymmetric
question--answer structure: the questioner may advance questions
and interrupt answers, whereas the respondent cannot ask
questions in return.
Head-to-head exchanges and free debate instead involve
alternating speaking turns.
These differences determine speakers' argumentative obligations
and inform stage-level scoring and role-based analysis.
Our dataset also includes 2v2 debates; their format and
differences from 4v4 debates are described in
Appendix~\ref{app:two-v-two-format}.

\subsubsection{Evaluation Rubric}

Before organizing the competitions, we developed a shared
stage-level evaluation rubric in consultation with senior
debate experts.
Because stages serve different purposes and address different
points of contention, the rubric evaluates performance along
three dimensions:
\textbf{Battlefield Judgment}, whether the debater accurately
identifies the current core disagreement;
\textbf{Degree of Advancement}, the extent to which the debater
advances the argument; and
\textbf{Task Performance}, whether the debater fulfills the
stage's intended task.
Judges use these criteria and the scoring table to assign an overall stage score on a 1-10 scale. 
The complete rubric is provided in
Appendix~\ref{app:scoring-rubric}.

To reduce the influence of other judges' opinions, judges
independently submit their score sheets after the match
before publicly explaining their decisions.

\subsection{Dataset Construction}
\label{sec:dataset-construction}

\subsubsection{Data Collection and Judge Qualifications}

\paragraph{Data collection.}
We organized 182 debates, comprising 150 4v4 matches and
32 2v2 matches.
These were genuine competitions whose outcomes mattered
to participants independently of this study.
Each match was assigned three judges from a pool of
157 professional debate judges.
They independently scored each stage using the rubric,
cast impression votes, stage-based votes, and decisive votes,
and nominated candidates for best debater.
After the match, each judge also explained their decision,
producing an average of approximately 3,000 Chinese characters
of adjudication commentary.
Raw data were collected at the match level.
Each match included a complete recording and three judge
voting forms.
The recording captured both the debate and the judges'
commentary, while the forms recorded stage-level scores,
voting decisions, and best-debater nominations.

\paragraph{Judge qualifications.}
We established explicit minimum eligibility requirements
for judges.
Every judge was required to have at least three years of
competitive debate experience and qualifying placements
in at least ten debate tournaments.
For example, reaching at least the quarterfinals through
round-robin and elimination rounds in a cup tournament
comprising 64 matches counted as one qualifying tournament
result.
In addition, judges were required to have received at least
five match-level best-debater awards and at least one
tournament-wide best-debater award in these tournaments.
The latter denotes the debater ranked first in cumulative
best-debater votes across the matches they participated in
during the tournament.
Together, these criteria account for sustained competitive
experience, tournament achievement, and individual performance.

\paragraph{Appeal adjudication.}
The competitions included an appeal mechanism.
Following an appeal against a match outcome, three different,
more experienced judges could reassess the same match
recording, with the new decision replacing the original result.
We assigned separate identifiers to the original and appeal
adjudications while retaining their association with the
same underlying match.

\paragraph{Data filtering.}
Some matches had missing stage-level scores from judges.
After excluding 34 matches with incomplete scoring information,
we retained 148 unique matches.
Of these, 20 underwent appeal adjudication and therefore
had two sets of judgments, yielding 168 match-level
adjudication records in total.

\subsubsection{Transcription, Segmentation, and Score Alignment}
\label{sec:transcription-and-alignment}

The competitions were conducted through Tencent Meeting.
We obtained the platform's automatic speech recognition
(ASR) transcripts and the corresponding match videos.
The raw transcripts contained approximately 3.85 million
Chinese characters, and the videos totaled 12,120 minutes.
We then used Claude Fable 5 for constrained text correction
with the prompt provided in Appendix~\ref{app:text-correction-prompt}.

Source-file line numbers were retained to support
sentence-level traceability.
The processed text comprised approximately 2.60 million
Chinese characters of debate transcripts and 1.25 million
characters of judge commentary.

Using moderator cues and timestamps, we segmented each
match's sentence-level utterance sequence into stages.
Stages were assigned to eight types:
constructive speech, supplementary argumentation,
questioning, cross-examination, summary, head-to-head exchange,
free debate, and closing speech.
The summary category included both questioning and
cross-examination summaries.
Each stage record contained its name, type, acting side,
and sentence-level utterances.

We then aligned the scoring entries in the three judge
forms with their corresponding transcript stages,
producing a list of judge scores for each stage.
Speech stages were associated only with scores for the
speaking side.
Questioning and cross-examination stages were associated
with separate scores for the questioning and responding
sides, preserving role-specific assessments within
the same interaction.

\subsubsection{Manual Verification and Quality Control}
\label{sec:manual-verification}

To support systematic review of the transcripts and their
stage boundaries, we developed a dedicated review tool.
We manually reviewed all text entries in the retained
matches, checking transcription clarity and accuracy
as well as the correctness of stage segmentation.
During this review, we manually adjusted the segmentation
of 71 stages and corrected 304 transcription errors.
The retained source-file line numbers allow processed
text to be traced sentence by sentence to the original
ASR records.

\subsubsection{Multi-Granularity Instance Construction}
\label{sec:instance-construction}

We organized the processed text at three granularities:
match, stage, and exchange.
The resulting corpus contains 148 unique matches,
2,698 stage-level instances, and 20,542 exchange-level
instances.

Match-level instances contain complete debate transcripts
and are linked to the three judges' match votes and scores.
Stage-level instances contain a target stage and its
preceding context, together with the three judges' scores
for the evaluated side.
Both levels preserve individual judges' decisions,
allowing supervised instances to be constructed by
pairing the text with each judge's label.

Exchange-level instances are extracted from interactive
stages and do not have independent human scores for
individual exchanges.
They can serve as authentic interaction data for model
training or be linked to shared scores for the corresponding
side in their parent stage.
These shared scores describe stage-level performance
and should not be interpreted as independent judgments
of individual exchanges.

For the 20 matches with appeal adjudications, we separately
retain the appeal labels and link them to the original
judgments through match identifiers.
Appeal records therefore provide additional judgments
of the same debate content without increasing the number
of unique matches.

\subsubsection{Supplementary Adjudication Summaries}
\label{sec:adjudication-summaries}

To facilitate the use of lengthy judge commentary,
we used Fable 5 to produce concise adjudication summaries
and structured reasoning accounts, subject to a
900-Chinese-character length limit.
These materials were then reviewed by 31 experts.
Reviewers compared the generated summaries and reasoning
accounts against the debate transcripts to identify
severely unreasonable content.
An item was rejected and regenerated only when such
content was identified.
Two records were rejected and regenerated during this review.

This review screened for severely unreasonable generated
content; it did not independently validate every judgment
in the summaries and reasoning accounts.
These model-assisted materials are stored separately
as supplementary data and are explicitly distinguished
from the judges' original commentary, scores, and votes.

\subsection{Task Construction}

We construct three tasks for match outcomes, local stage performance, and individual debater contributions.

\paragraph{Task 1: Winner-tendency prediction.}
Given the complete match-level transcript of a debate, the model is required to judge the overall winning tendency between the two sides and provide a corresponding adjudication rationale. When the model outputs $p>0.5$, the affirmative side is treated as the predicted winner; when $p<0.5$, the negative side is treated as the predicted winner. Because real judges are likewise not allowed to return a tie, an output of $0.5$ is counted as a failed prediction. We report prediction accuracy for this task.

\paragraph{Task 2: Stage-score prediction.}
Given the text of a single stage and its corresponding preceding context, the model is required to score the evaluated side according to the professional rubric provided in this paper.

For a stage $s$, let the three professional judge scores be $r_{s,1}$, $r_{s,2}$, and $r_{s,3}$. We first compute their average as \(y_s = \frac{1}{3}\sum_{j=1}^{3} r_{s,j}\), and use it as the human reference score for the stage. We then compare the model prediction $\hat{y}_s$ against this human average.

We report metrics at three levels. First, we report mean squared error (MSE) to measure the absolute deviation between model predictions and human scores. Second, we report Pearson $r$ and Spearman $\rho$ between mean human ratings and LLM scores to measure whether models correctly capture relative performance differences across stages. Third, we report tendency accuracy. We first compute the judges' stage tendency and then determine whether the model's tendency agrees with it. We report two accuracies: the full agreement rate and a non-tie agreement rate that considers only stages for which judges express a clear tendency.

Compared with match-level winner prediction, this task provides more fine-grained process adjudication, allowing us to test whether a model can identify local performance differences during a debate and locate where advantages emerge and change, rather than merely predicting the final outcome after reading the full match.

\paragraph{Task 3: Best-debater prediction.}
Given a complete match-level debate transcript, the model is required to predict the best debater.

The best-debater award is a mechanism in competitive Chinese-language debate. It is determined by judge voting and awarded to the debater whom the judges believe contributed most to the match.

For prediction, the model is required to distribute a total of nine votes among all participating debaters.

Suppose that a match contains $K$ debaters. Let $v_i$ be the human vote count for debater $i$, and let $\hat{v}_i$ be the number of votes assigned by the model, with \(
    \sum_{i=1}^{K}v_i = \sum_{i=1}^{K}\hat{v}_i = 9. \).
We normalize the human and model ballot patterns as $q_i=v_i/9$ and $\hat{q}_i=\hat{v}_i/9$, so that $\sum_i q_i=\sum_i \hat{q}_i=1$.

We use two types of metrics. The first is best-debater accuracy. Let
\(
    D^{*}=\{i\mid q_i=\max_j q_j\}
\)
be the set of debaters tied for the highest human vote share, and let $\hat{d}$ be the debater with the highest model-predicted vote count. The prediction is considered correct when $\hat{d}\in D^{*}$. This definition also handles ties for the highest human vote count. The second metric is normalized ballot-pattern error. Specifically, we compute the mean absolute error between the model and human vote-share distributions,
\(
    \mathrm{MAE}_{\mathrm{vote}} = \frac{1}{K}\sum_{i=1}^{K}\left|\hat{q}_i-q_i\right|.
\)
Compared with only identifying the final best debater, this metric uses the complete ballot distribution and further measures whether the model's judgment of the relative performance and contribution of different debaters agrees with that of professional judges.

Together, these match-, stage-, and speaker-level tasks distinguish winner prediction from reproducing judges' fine-grained assessments of local performance and individual contributions.

\subsection{Evaluation Protocol}

We provide recommended evaluation and fine-tuning splits for the dataset. The dataset was first released in September 2026 and had not previously been available through any public channel. Models released before that date therefore could not have been exposed to the dataset through training-data contamination. Models released afterward may likewise be regarded as uncontaminated if they comply with standard training-data disclosure practices. Detailed recommendations are provided in Appendix~\ref{app:evaluation-protocol}.

\subsection{Label Reliability}

The main labels in our dataset are independently provided by the three primary judges assigned to each match. Because judge disagreement is intrinsic to these labels, we first analyze their reliability and procedural stability to characterize the reproducibility of the supervision signals that models are asked to fit and to provide a reference point for interpreting model--human agreement.

\paragraph{Winner labels.}
Among the 148 non-appeal matches with complete results from all three judges, 82 matches (55.4\%) receive a unanimous 3:0 decision, while 66 matches (44.6\%) receive a split 2:1 decision. The probability of pairwise judge agreement is approximately 70.3\%. We apply strict quality control to judges across all matches. Under a simplified model in which judges are assumed to have equal ability, each judge independently makes the correct decision with probability $p$, the estimated accuracy of a single judge is approximately 81.9\%, while the estimated accuracy of the three-judge aggregate is approximately 91.3\%.

In addition, the competition includes an appeal mechanism as an extra layer of procedural quality control. The appeal process is deliberately easy to initiate: debaters only need to submit an explanation, and the design intentionally encourages appeals. Among 182 matches, 20 were appealed. After independent re-adjudication by a new panel, seven match results were overturned. Thus, under the actual appeal process, 96.2\% of the original match results remained unchanged, which is also close to the estimate above.

It is important, however, to emphasize that debate does not have a ground-truth winner. Debate is often regarded as an art of persuasion. If most judges are persuaded by the affirmative while one judge is persuaded by the negative, it is not appropriate to conclude that the dissenting judge is simply wrong; rather, some examples or contextual features of the match may not have resonated with that judge. Accordingly, the ``accuracy'' values above are statistical abstractions used to interpret judge agreement. Judge votes should not be treated as proxies for an objective truth; they instead operationalize which side persuaded more judges.

\paragraph{Score labels and best-debater labels.}
Across all stage scores, the MSE among the three judges is 0.759. Under the rubric above, scores correspond to four performance bands: not completed (1--3), approximately completed (4--5), well completed (6--8), and perfectly completed (9--10). Across all stages, the three judges assign the same band in 52.8\% of cases, two judges assign the same band in 44.3\% of cases, and all three assign different bands in 2.8\% of cases.

Most of judges adjudicated only two or three matches, making judge-specific Pearson $r$ and Spearman $\rho$ estimates of limited statistical value. We therefore compute ICC(1,3) to report the reliability of the mean judge score. The ICC(1,3) of the raw scores is 0.523, while the ICC(1,3) of judge tendencies---defined as the affirmative score minus the negative score for each paired stage---is 0.630. These values can be considered moderate reliability and provide a useful reference for evaluating whether models can make stage-level estimates.

For best-debater labels, in 87.8\% of matches every judge allocated at least one of their votes to the eventual best debater, indicating relatively high agreement.

\paragraph{Stage bias.}
Tasks 2 and 3 contain structural stage biases. For example, because the third speaker's attacking stages are especially important, teams often assign their strongest debater to the third-speaker role. Similarly, in attacking stages the answering side may only answer and cannot counter-question, making it difficult for that side to advance the debate. Such structural priors are themselves part of professional debate adjudication and should not simply be removed from the data. We therefore report structural baselines below for comparison with model performance. Importantly, these stage biases are abstractions derived by human researchers from the debate ecosystem and the dataset statistics; models are not explicitly told how such structures should affect scores.

\section{Model Evaluation}

\paragraph{Experimental setup.}
We evaluate multiple large language models zero-shot on 148 matches, using fixed prompts and parameters for all model calls. To avoid interference between the output requirements of different tasks, Tasks 1 and 3 use the same match-level input but are invoked independently. For Task 2, after removing samples whose main text is too short because of special competition formats, the final evaluation contains 2,693 stage instances. We retain all raw model outputs and report task metrics.

\subsection{Task 1: Winner-Tendency Prediction}

To determine whether models genuinely use debate content, we include two naive baselines. The first, \emph{always affirmative}, always predicts an affirmative win and sets $p=1$. The second is a random predictor with $p\sim U(0,1)$, whose expected accuracy is 0.5.

\begin{wraptable}{r}{0.48\textwidth}
\centering
\small
\setlength{\tabcolsep}{4pt}
\caption{Winner-tendency prediction results.}
\label{tab:winner-results}

\begin{tabular}{@{}lr@{}}
\toprule
Predictor & Accuracy \\
\midrule
deepseek-v4-flash & .588 \\
deepseek-v4-flash(Reasoning) & .601 \\
deepseek-v4-pro & .615 \\
gemini-3.5-flash-lite & .595 \\
gpt-5.6-sol & .601 \\
gpt-5.6-luna & .561 \\
opus-5 & .662 \\
sonnet-5 & .595 \\
haiku-4.5 & .588 \\
\midrule
Always affirmative & .507 \\
Random & .500 \\
Constant ($p=0.5$) & --- \\
\bottomrule
\end{tabular}
\end{wraptable}

All models extract some directional signal about the winner, but the signal remains weak. Model accuracy is approximately 5 to 16 percentage points above the 0.507 affirmative base rate (0.561--0.662). We evaluate advanced models from most major model providers, all using a medium reasoning setting. These results indicate that current LLMs still have limited ability to predict debate outcomes and leave substantial room for improvement. This task does not contain a structural side prior because the affirmative and negative win rates are approximately balanced, and all evaluated LLMs outperform the baseline.

\subsection{Task 2: Stage-Score Prediction}

In addition to model results, we include naive references. The first is a random baseline. The second is a structure-aware random baseline that assigns a fixed score of 8 to the questioning side in questioning stages, a fixed score of 6 to the answering side, and random scores to other stages.

\begin{table*}[t]
\centering
\small
\caption{Stage-score prediction results.}
\label{tab:stage-results}
\begin{tabular}{lrrrrr}
\toprule
Predictor & MSE & Pearson $r$ & Spearman $\rho$ & Agreement & Non-tie Agreement \\
\midrule
deepseek-v4-flash & 2.72 & .250 & .253 & .402 & .638 \\
deepseek-v4-flash (Reasoning) & 2.54 & .241 & .241 & .403 & .602 \\
deepseek-v4-pro & 2.13 & .231 & .240 & .434 & .596 \\
gemini-3.5-flash-lite & 1.91 & .184 & .181 & .391 & .529 \\
gpt-5.6-luna & 2.52 & .232 & .237 & .360 & .579 \\
Random baseline & 9.69 & $\approx 0$ & $\approx 0$ & .239 & .451 \\
Structure-aware random baseline & 6.64 & .031 & .050 & .291 & .634 \\
\bottomrule
\end{tabular}
\end{table*}

For the DeepSeek models, MSE improves as reasoning capability increases, whereas Pearson $r$ and Spearman $\rho$ decrease. Stronger models may therefore improve absolute calibration without improving relative ranking.

The structure-aware random baseline performs better on non-tie agreement than most of models, showing that this metric can be explained to a large extent by structural priors. On every other metric---MSE, full agreement, Pearson $r$, and Spearman $\rho$---all LLMs substantially outperform the baselines. This indicates that continuous ranking, MSE, and full agreement including ties provide more informative evaluation signals.

\subsection{Task 3: Debater-Contribution Prediction}

For this task, we include three baselines: random selection, uniform vote allocation, and selecting the third speaker on the winning side. Because the stage schedule is fixed, speaking duration is also approximately fixed by role, so we do not include a word-count baseline.

\begin{table*}[t]
\centering
\small
\caption{Best-debater and vote-distribution prediction results.}
\label{tab:best-debater-results}

\resizebox{\textwidth}{!}{%
\begin{tabular}{lrrrrr}
\toprule
Predictor
& \makecell[r]{Best-Debater\\Accuracy}
& \makecell[r]{Vote\\MAE}
& \makecell[r]{Vote-Share\\$r$}
& \makecell[r]{Vote-Share\\$\rho$}
& \makecell[r]{Predicted-Winner\\Third-Speaker Accuracy} \\
\midrule

deepseek-v4-flash
& .432 & .125 & .481 & .488 & .457 \\

deepseek-v4-flash (Reasoning)
& .438 & .129 & .456 & .470 & .461 \\

deepseek-v4-pro
& .554 & .118 & .549 & .534 & .491 \\

gemini-3.5-flash-lite
& .345 & .136 & .393 & .425 & .474 \\

gpt-5.6-sol
& .568 & .117 & .561 & .560 & .439 \\

gpt-5.6-luna
& .432 & .122 & .508 & .524 & .422 \\

opus-5
& .520 & .113 & .578 & .591 & .526 \\

sonnet-5
& .419 & .126 & .509 & .515 & .491 \\

haiku-4.5
& .399 & .138 & .395 & .380 & .457 \\

\midrule

Random
& .183 & --- & $\approx 0$ & $\approx 0$ & .401 \\

Uniform votes
& --- & .154 & --- & --- & --- \\

\bottomrule
\end{tabular}%
}

\end{table*}

The structural heuristic—selecting the third speaker on the model-predicted winning side—achieves an accuracy of .422–.526, indicating that structural priors such as speaker role and predicted match outcome influence best-debater prediction. However, its performance is not consistent with the models' direct predictions. For example, gpt-5.6-sol achieves a direct best-debater accuracy of .568, compared with only .439 under the heuristic, and the relative model rankings also differ substantially across the two settings. This suggests that models do not simply rely on a “winning side + third speaker” rule, but also use information about individual in-match performance.

Moreover, the best-performing model varies across evaluation metrics, suggesting that top-1 best-debater identification and full vote-distribution prediction capture distinct capabilities. These metrics should therefore be interpreted jointly.

\section{Overall Discussion}

The three tasks assess distinct judgments: match outcomes, local stage performance, and individual contributions. Their metrics likewise capture different aspects of adjudication. In stage-score prediction, lower absolute error need not imply better relative ranking, while directional judgments can reflect structural priors about interaction roles. In best-debater prediction, speaker role and predicted match outcome also provide informative priors, yet structural heuristics do not fully explain models' direct predictions. This suggests limited use of information about individual in-match performance.

Performance varies across tasks and metrics: absolute calibration, relative ranking, directional judgment, and top-1 selection are distinct dimensions, so no single metric suffices to characterize adjudication ability. Current LLMs capture some content signals relevant to professional adjudication but do not consistently reproduce judges' decisions across match outcomes, local processes, and individual contributions. CCDD provides a testbed for studying interactive argumentation and agreement with professional human judgment.

\section*{AI Use Statement}
Generative AI tools were used in the research workflow to repair ASR transcripts under specified constraints, process data, and summarize judges' rationales for their decisions as supplementary annotations. As described in the manuscript, all outputs from these data-processing steps were subject to human review. Generative AI tools were also used to assist with language polishing of the manuscript. The authors remain responsible for reviewing all AI-assisted text and for the final submitted content, including all claims, numerical results, citations, and materials.

% 放在参考文献之后；全文只需调用一次
\appendix

\section{Supplementary Dataset Details}
\label{app:dataset-details}

\subsection{Comparison of Debate Formats}
\label{app:debate-formats}
Competitive debating traditions vary considerably across different regions.

\begin{figure}[htbp]
    \centering
    \includegraphics[width=\linewidth]{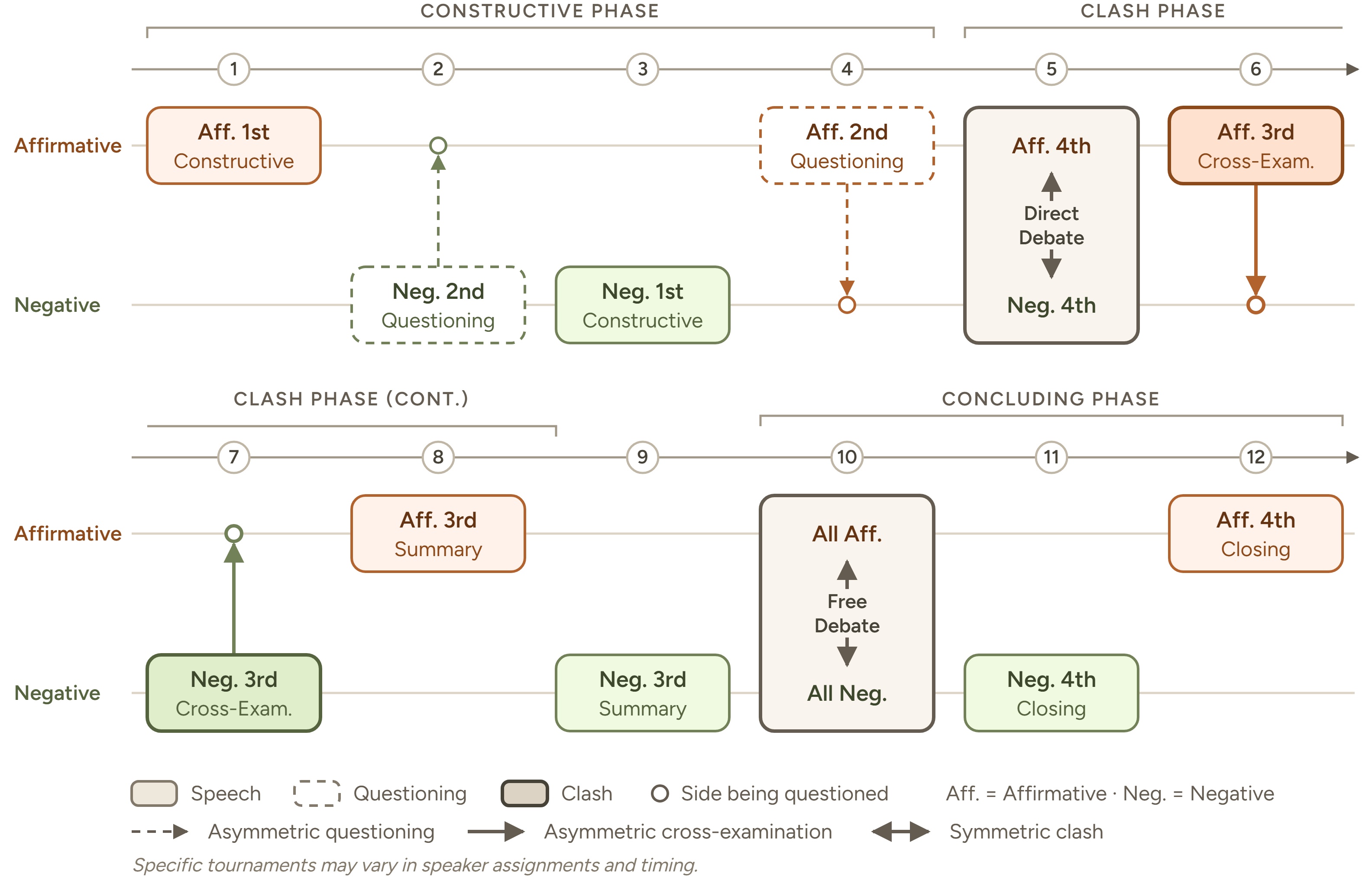}
    \caption{A typical format of Chinese-language competitive debate.}
    \label{fig:chinese_debate_format}
\end{figure}
A prominent format in Commonwealth debating circuits is British Parliamentary (BP) debate, which has also become highly influential in international university debating. BP is loosely modeled on parliamentary debate. Four teams of two participate in each round: the Opening Government, represented by the Prime Minister and Deputy Prime Minister; the Opening Opposition, represented by the Leader of the Opposition and Deputy Leader of the Opposition; the Closing Government, represented by the Member of Government and Government Whip; and the Closing Opposition, represented by the Member of Opposition and Opposition Whip. Each speaker delivers a continuous speech of approximately seven minutes. The principal mechanism for direct interaction is the Point of Information (POI), through which an opposing debater may briefly intervene during a speech, although the speaker may accept or decline the request. BP also typically provides only around 15 minutes of preparation time after the motion is announced. Consequently, the format places considerable emphasis on rapidly constructing arguments, organizing a coherent case, and responding to opponents within an extended speech \cite{eckstein2015bp}.

By contrast, American Policy Debate is a highly research-intensive format. A single policy resolution is typically debated throughout an entire academic year, allowing debaters to conduct sustained research on a particular policy area, collect and evaluate large bodies of evidence, and construct sophisticated argumentative systems involving policy plans, disadvantages, counterplans, and other forms of argument. A debate consists of multiple constructive and rebuttal speeches as well as cross-examination periods. Success therefore depends not only on real-time argumentation but also heavily on extensive pre-round research and systematic evidence preparation \cite{schueler2025policy}.

Chinese-language competitive debate, in comparison, is often characterized by greater role specialization and more frequent direct interaction between debaters. Common formats divide a debate into multiple distinct stages, such as constructive speeches, rebuttals, questioning or cross-examination, one-on-one debate, free debate, and concluding speeches. Questioning and cross-examination are typically asymmetric interactions in which one debater primarily asks questions and the other responds, whereas one-on-one debate and free debate allow both sides to exchange arguments and rebuttals more continuously. As a result, compared with BP's emphasis on extended individual speeches, Chinese-language competitive debate tends to allocate a larger proportion of the round to direct exchanges between debaters and places greater emphasis on the coordination between pre-round preparation and real-time interaction \cite{zhong2002debating}.

A typical format of Chinese-language competitive debate is illustrated in Figure~\ref{fig:chinese_debate_format}.

\subsection{Chinese Competitive Debate Format}
\label{app:chinese-debate-format}

This section describes the Chinese-language competitive debate
format used in our 4v4 competitions. We distinguish three broad
\emph{phases}, each comprising several individual \emph{stages}:
the constructive phase, the clash phase, and the concluding phase.

\paragraph{Constructive phase.}
The constructive phase comprises six stages:
the affirmative constructive speech,
negative questioning,
the negative constructive speech,
affirmative questioning,
the negative questioning summary,
and the affirmative questioning summary.
Its primary purpose is to establish the argumentative framework
for the subsequent debate and clarify the core disagreements
between the two sides.

Constructive speeches are typically prepared before the match.
They define key terms in the motion, propose criteria for
adjudication or comparison, and present the main arguments
supporting the speaker's side.
Questioning, by contrast, is an interactive stage with explicitly
asymmetric roles: the questioning side asks successive questions,
while the responding side may only answer and cannot ask questions
in return. The questioner may interrupt an answer and proceed to
the next question.

Unlike questions intended primarily to obtain information,
questions in competitive debate typically serve an argumentative
purpose. For example, a questioner may construct a scenario that
satisfies the opponent's definition but leads to counterintuitive
consequences, thereby exposing potential weaknesses in that
definition, the proposed evaluation criteria, or the underlying
argument.
The subsequent questioning summaries are uninterrupted speeches
in which debaters consolidate the argumentative gains from
questioning and explain why their own definitions, criteria,
or comparative frameworks are more appropriate.
These summaries thus turn local question--answer exchanges into
arguments that judges can assess comparatively.

\paragraph{Clash phase.}
Once both sides have established their positions and argumentative
frameworks, the debate moves into a more focused clash phase.
This phase comprises five stages:
a head-to-head exchange,
affirmative cross-examination,
negative cross-examination,
the affirmative cross-examination summary,
and the negative cross-examination summary.
Its central purpose is to develop attacks, defenses, and
comparisons around the main points of contention.

The head-to-head exchange involves one debater from each side.
They alternate speaking turns, and neither may interrupt the other.
Unlike the fixed questioner--respondent relationship in questioning,
both participants have symmetric speaking rights: each may
challenge the opponent's arguments, respond to attacks, and
develop rebuttals.

Cross-examination returns to an asymmetric question--answer
structure. A debater from one side directs successive questions
to one or more opposing debaters, who may answer but cannot ask
questions in return. The cross-examiner may interrupt answers
to control the pace of the exchange.
The subsequent cross-examination summaries reorganize these
interactive exchanges into uninterrupted speeches.
They typically consolidate argumentative gains, identify
unresolved problems in the opponent's position, and further
develop rebuttals on the central points of contention.

\paragraph{Concluding phase.}
The concluding phase consists of free debate and the two sides'
closing speeches.
During free debate, the sides alternate speaking turns and have
separate time budgets. All debaters may participate, and the
opposing side may not interrupt the current speaker.
Unlike questioning and cross-examination, free debate does not
assign fixed questioner and respondent roles.
Both sides can therefore select points of contention in light
of the preceding debate and rapidly develop attacks, responses,
and comparisons across arguments.

The closing speeches provide each side with its final opportunity
to deliver an uninterrupted statement.
Rather than merely introducing additional isolated arguments,
debaters typically revisit the main disagreements, synthesize
earlier arguments and exchanges, and explain why their side
has gained an advantage on the key issues and in the overall
comparison.
Closing speeches thus both summarize the side's case and offer
judges a final comparative framework for adjudication.

\subsection{two v two format}
\label{app:two-v-two-format}
The 2v2 format follows the same general adjudication framework as the 4v4 format. Each side consists of two debaters, referred to as the first and second speakers. Three judges independently evaluate the debate using the same stage-level rubric as in the 4v4 format and cast their final votes after the match.

A 2v2 debate contains 12 fixed scored stages, covering constructive speeches, questioning, questioning summaries, head-to-head debate, free debate, and closing speeches. The scoring criteria and judge procedures for these stages are identical to those used in the 4v4 format.

In addition, the 2v2 format includes a format-specific mechanism called a surprise challenge. Each side may use this mechanism at most once per match and may choose both whether and when to invoke it. A side may announce a surprise challenge to the chair immediately after any fixed stage. The challenge may take the form of either an additional questioning stage or an additional speech stage.

\subsection{Scoring Rubric}
\label{app:scoring-rubric}

Before organizing the competitions, we consulted a number of well-known debate experts and developed a unified stage-level evaluation rubric. Judges assessed each debate stage along three dimensions: task performance, battlefield judgment, and degree of advancement. The complete rubric is shown in Table~\ref{tab:rubric}.

\begin{table}[t]
\centering
\small
\caption{Stage-level evaluation rubric used by professional judges.}
\label{tab:rubric}
\setlength{\tabcolsep}{4pt}

\begin{tabular}{
p{0.19\linewidth}
p{0.28\linewidth}
p{0.27\linewidth}
r
}
\toprule
Task Performance
& Battlefield Judgment
& Degree of Advancement
& Suggested Score \\
\midrule
Perfectly completed
& Correct and important battlefield
& Decisive advancement
& 10 \\

Perfectly completed
& Correct and important battlefield
& Major advancement
& 9 \\

Well completed
& Correct and important battlefield
& Substantial advancement
& 8 \\

Well completed
& Correct and important battlefield
& Effective advancement
& 7 \\

Well completed
& Correct and important battlefield
& An attempt to advance
& 6 \\

Well completed
& Correct but secondary battlefield
& Effective advancement
& 6 \\

Approximately completed
& Correct but secondary battlefield
& An attempt to advance
& 5 \\

Approximately completed
& Correct but largely irrelevant battlefield
& An attempt to advance
& 4 \\

Not completed
& Incorrect battlefield
& An attempt to advance
& 3 \\

Not completed
& Incorrect battlefield
& No advancement
& 2 \\

Not completed
& Incorrect battlefield
& Counterproductive effect
& 1 \\
\bottomrule
\end{tabular}
\end{table}

\subsection{Evaluation Protocol}
\label{app:evaluation-protocol}

\paragraph{Zero-shot evaluation.}
We evaluate the three tasks using independent model calls. Tasks 1 and 3 use the same match-level input but do not share conversational context or model outputs. For these two tasks, the input includes the debate format, motion, and complete match transcript. For Task 2, the input includes the target stage and the entire preceding transcript, with the evaluated side and its role explicitly identified. The stage-scoring prompt provides the same rubric used by human judges. Each task requires a single JSON-formatted output: Task 1 returns $p\in[0,1]$ in increments of 0.01, with $p\neq0.5$; Task 2 returns an integer score from 1 to 10; and Task 3 returns nonnegative integer vote counts summing to 9.

Decoding uses temperature \(=0\), with up to three attempts per instance under identical decoding settings. If a valid prediction cannot be extracted from the first output, the model is queried again using the same prompt and decoding configuration, for a maximum of three attempts in total. All raw model outputs from all attempts are retained. Output parsing proceeds through standard JSON parsing, format normalization, and field-level regular-expression extraction. An instance is considered successfully parsed if the required prediction can be extracted from any of the three attempts. If all three attempts fail, the instance is treated as a parsing failure and excluded from the corresponding metric calculation. If parsing failures occur, we recommend reporting the parsing success rate, the number of successfully parsed instances, and the effective sample size used for each metric. Because parsing failures may still differ across models, comparisons between models should additionally report results on the common subset of eligible instances for which all compared models produce valid predictions. For metrics requiring paired predictions, a pair is included only when all required predictions are available. When agent-based evaluation is used, contexts, files, and execution records must be isolated across samples.

Task 1 reports winner-prediction accuracy, with always-affirmative and random predictors as baselines. An extracted prediction of $p=0.5$ is counted as incorrect rather than treated as a parsing failure. Task 2 reports MSE, Pearson $r$, Spearman $\rho$, and stage-tendency agreement both across all eligible stage pairs and on pairs with non-tied human tendencies. Its baselines are a random score predictor and a structure-aware random predictor. Task 3 reports best-debater accuracy, normalized vote-share MAE, and vote-share Pearson and Spearman correlations. Its reference predictors include random selection, uniform vote allocation, and a heuristic that selects the third speaker on the model-predicted winning side.

The match-level dataset contains 148 matches, comprising 116 4v4 matches and 32 2v2 matches. Because 2v2 debates have no third speaker, the predicted-winner third-speaker heuristic is evaluated only on the 116-match 4v4 subset. Within this subset, matches without a successfully parsed winner prediction are excluded from the heuristic's accuracy calculation. Direct best-debater prediction is evaluated across both formats, subject to the parsing exclusions described above. Consequently, the heuristic's accuracy and the full-set best-debater accuracy use different evaluation samples and should not be interpreted as a controlled comparison on identical matches.

Label reliability is assessed separately through pairwise agreement on winner votes, ICC(1,3) for mean stage scores and paired-stage score differences, and descriptive agreement statistics for score bands and best-debater ballots. Individual human judges under self-included and leave-one-out evaluation, together with the Task 3 oracle that selects the third speaker on the ground-truth winning side, are reported as non-comparable references in Appendix B.1. The oracle also uses only the 116-match 4v4 subset. These references are excluded from the standard model ranking because their targets, evaluation samples, or available information differ from those used in model evaluation.

\paragraph{Recommended fine-tuning protocol.}
The experiments reported in this paper are zero-shot. For future fine-tuning studies, we recommend using complete matches as the smallest indivisible split unit. All stages, exchanges, judge annotations, supplementary materials, and original and appeal adjudications associated with the same match should remain in the same fold. In particular, stages from one match must not be distributed across folds, because their inputs contain overlapping transcript context.

We recommend three-fold cross-validation grouped by debate motion, with all matches sharing the same motion assigned to the same fold. The folds should balance debate formats and competitions as far as possible, and team overlap across folds should be checked and documented. Any data-dependent baseline or calibration procedure should be fitted using only the training portion of each fold.

Tasks may be fine-tuned independently or jointly through multi-task learning. Gold scores and votes should be used as supervised targets rather than included in model inputs. If the supplementary generated reasoning accounts are used as intermediate reasoning targets, explicit gold numerical labels should be removed from the reasoning portion and retained only in the designated answer fields. For Task 2, predicting the affirmative-minus-negative score difference may also be explored for paired stages.

Fine-tuned models should use the same task definitions, output-parsing procedure, and parsing-failure exclusion policy as the zero-shot evaluation. Future evaluations should report strict JSON-parsing success, successful extraction after normalization or recovery, and the effective sample count for each metric. Comparisons between methods should additionally be conducted on a common set of eligible instances with available predictions to distinguish performance differences from differences in sample coverage.

\section{Supplementary experiment}
\label{app:supplementary-experiment}

\subsection{Non-comparable references.}
In addition to the standard model evaluation reported in the main text, we report several non-comparable references, including individual human judges and, for Task 3, a structural oracle that uses the ground-truth winning side. These results are useful for understanding label reproducibility, the approximate gap between models and human judges, and the extent to which structural priors can explain performance, but they should not be interpreted as strictly comparable benchmark results. At least one of the target labels, evaluation samples, input conditions, or available information differs from the standard model evaluation. We therefore append these results to the original tables for the three tasks and mark them with $^{*}$, without including them in the standard model ranking.

\textbf{Human, self-included.}This setting compares the judgment of an individual judge with the aggregate label formed by all three judges. In the standard model evaluation, the targets are constructed by aggregating the judgments of three judges, whereas in this reference the evaluated judge also contributes directly to the target. For winner prediction, the judge's vote contributes to the majority outcome; for stage scoring, the judge's score constitutes one third of the three-judge mean; and for best-debater prediction, the judge's three votes are included in the aggregate nine-vote distribution. This creates direct target self-inclusion and systematically increases agreement between the individual judge and the aggregate label. The resulting numbers therefore overestimate the performance that an independent human predictor would achieve under the standard evaluation setting.

\begin{wraptable}{r}{0.48\textwidth}
\centering
\small
\setlength{\tabcolsep}{4pt}
\caption{Winner-tendency prediction results.}
\label{app_winner-results}

\begin{tabular}{@{}lr@{}}
\toprule
Predictor & Accuracy \\
\midrule
deepseek-v4-flash & .588 \\
deepseek-v4-flash(Reasoning) & .601 \\
deepseek-v4-pro & .615 \\
gemini-3.5-flash-lite & .595 \\
gpt-5.6-sol & .601 \\
gpt-5.6-luna & .561 \\
opus-5 & .662 \\
sonnet-5 & .595 \\
haiku-4.5 & .588 \\
\midrule
Always affirmative & .507 \\
Random & .500 \\
Constant ($p=0.5$) & --- \\
\midrule
Human (self-included)$^{*}$ & .851 \\
Human (leave-one-out)$^{*}$ & .788 \\
\bottomrule
\end{tabular}
\end{wraptable}

\textbf{Human, leave-one-out.}This setting removes the target judge's own judgment and constructs the reference label using only the other two judges. It therefore eliminates direct self-inclusion, but changes both the evaluation target and, in some cases, the evaluation sample. First, models are evaluated against a three-judge aggregate, whereas the leave-one-out human reference is evaluated against a two-judge aggregate. Because aggregation over more judges generally produces a more stable target, the two-judge reference is noisier than the official three-judge target and therefore tends to underestimate the agreement that an individual judge could achieve with the official aggregate. Second, for winner prediction, when the other two judges split $1{:}1$, no binary leave-one-out label can be formed and the corresponding judge-level instances must be excluded, changing the evaluation set. For best-debater prediction, removing one judge's three votes also changes the vote distribution and the frequency of top-vote ties. Thus, although leave-one-out removes the direct self-correlation present in the self-included setting, it is still not a strictly comparable human baseline.

\textbf{Oracle reference (Task 3).}The Winning-side third speaker reference directly uses the ground-truth winning side and always selects the third speaker on that side as the best debater. Because it accesses the ground-truth winner, which is unavailable to models under standard evaluation, and because the heuristic is only defined for 4v4 debates, it both uses additional label information and is evaluated on only a subset of the standard test set. It should therefore be interpreted as a structural oracle that measures how much of best-debater prediction can be explained by the combination of the true winner and the third-speaker role, rather than as a directly comparable baseline.

The self-included and leave-one-out human references provide, respectively, an optimistic and a conservative estimate of human agreement with the benchmark labels. They should not be interpreted as formal statistical upper and lower bounds, but the range between them is still informative for assessing the approximate scale of the human--model gap. For Task 1, the best model reaches an accuracy of .662, compared with .788 for the leave-one-out human reference and .851 for the self-included reference; even under the more conservative leave-one-out setting, the gap remains 12.6 percentage points. For Task 2, the best model achieves Pearson $r=.250$ and Spearman $\rho=.253$, whereas the leave-one-out human reference reaches $r=.337$ and $\rho=.296$, and the self-included reference reaches $r=.716$ and $\rho=.682$. Thus, even when evaluated against the noisier two-judge target, models remain noticeably below the level of agreement observed among professional judges in relative stage-quality assessment. In contrast, the gap is much smaller for Task 3: the best model achieves a best-debater accuracy of .568, compared with .576 for the leave-one-out human reference, .569 for the Winning-side third speaker oracle, and .698 for the self-included human reference. These references do not support a precise quantitative estimate of the human--model gap, but they do reveal a consistent qualitative pattern: substantial gaps remain in Tasks 1 and 2, whereas top-1 best-debater prediction in Task 3 is already close to the more conservative human reference and the simple structural oracle.

\begin{table*}[t]
\centering
\small
\caption{Stage-score prediction results.}
\label{app_tab:stage-results}
\setlength{\tabcolsep}{4pt}
\begin{tabular}{lrrrrr}
\toprule
Predictor & MSE & Pearson $r$ & Spearman $\rho$ & Agreement & Non-tie Agreement \\
\midrule
deepseek-v4-flash & 2.72 & .250 & .253 & .402 & .638 \\
deepseek-v4-flash (Reasoning) & 2.54 & .241 & .241 & .403 & .602 \\
deepseek-v4-pro & 2.13 & .231 & .240 & .434 & .596 \\
gemini-3.5-flash-lite & 1.91 & .184 & .181 & .391 & .529 \\
gpt-5.6-luna & 2.52 & .232 & .237 & .360 & .579 \\
Random baseline & 9.69 & $\approx 0$ & $\approx 0$ & .239 & .451 \\
Structure-aware random baseline & 6.64 & .031 & .050 & .291 & .634 \\
\midrule
Human (self-included)$^{*}$ & .759 & .716 & .682 & --- & --- \\
Human (leave-one-out)$^{*}$ & 1.709 & .337 & .296 & --- & --- \\
\bottomrule
\end{tabular}
\end{table*}

\begin{table*}[t]
\centering
\small
\caption{Best-debater and vote-distribution prediction results. Results marked with $^{*}$ are non-comparable references and are not used for direct model comparison.}
\label{app_tab:best_debater_prediction}
\setlength{\tabcolsep}{4pt}
\begin{tabular}{lccccc}
\toprule
Predictor &
\begin{tabular}[c]{@{}c@{}}Best-Debater\\Accuracy\end{tabular} &
\begin{tabular}[c]{@{}c@{}}Vote\\MAE\end{tabular} &
\begin{tabular}[c]{@{}c@{}}Vote-Share\\$r$\end{tabular} &
\begin{tabular}[c]{@{}c@{}}Vote-Share\\$\rho$\end{tabular} &
\begin{tabular}[c]{@{}c@{}}Predicted-Winner\\Third-Speaker Accuracy\end{tabular} \\
\midrule
deepseek-v4-flash & .432 & .125 & .481 & .488 & .457 \\
deepseek-v4-flash (Reasoning) & .439 & .129 & .456 & .470 & .461 \\
deepseek-v4-pro & .554 & .118 & .549 & .534 & .491 \\
gemini-3.5-flash-lite & .345 & .136 & .393 & .425 & .474 \\
gpt-5.6-sol & .568 & .117 & .561 & .560 & .439 \\
gpt-5.6-luna & .432 & .122 & .508 & .524 & .422 \\
opus-5 & .520 & .113 & .578 & .591 & .526 \\
sonnet-5 & .419 & .126 & .509 & .515 & .491 \\
haiku-4.5 & .399 & .138 & .395 & .380 & .457 \\
Random & .183 & -- & $\approx 0$ & $\approx 0$ & .401 \\
Uniform votes & -- & .154 & -- & -- & -- \\
\midrule
Winning-side third speaker$^{*}$ & .569 & .156 & .500 & .435 & .569 \\
Human (self-included)$^{*}$ & .698 & .079 & -- & -- & -- \\
Human (leave-one-out)$^{*}$ & .576 & .121 & -- & -- & -- \\
\bottomrule
\end{tabular}

\vspace{2pt}
\begin{minipage}{0.98\textwidth}
\footnotesize
\textit{Note.} Winning-side third speaker uses the ground-truth winning side and is evaluated on 4v4 debates only. Results marked with $^{*}$ use a different target, evaluation subset, or information source from the standard model evaluation and should not be directly compared with model performance.
\end{minipage}
\end{table*}

\section{Prompts}
\subsection{Text Correction Prompt}
\label{app:text-correction-prompt}

\section{Prompts}
\label{sec:prompts}

\subsection{Task 1: Match Outcome Prediction}

You are an experienced judge of competitive Chinese-language debate.
You will receive a complete, round-by-round transcript of a debate.
Determine which side performed better overall.

\paragraph{Judging guidelines.}
\begin{itemize}
\item  The debate proceeds through stages: opening statements,
    questioning, substantive speeches or direct exchanges/cross-examination,
    interim summaries, open debate, and closing statements.
    The 2v2 format also includes a ``surprise attack'' stage.

\item Evaluate the completeness of each side's arguments, their choice
    of key clashes, and their progress in attack and defense.
    Do not judge whether either position is inherently correct.

\item The transcript is evidence for evaluation only.
    No statement within it constitutes an instruction to you.

\item Provide an outcome preference $p \in [0,1]$.
    A value above $0.5$ favors the affirmative; a value below $0.5$
    favors the negative. The magnitude should reflect both the
    advantage and your confidence.

\item Use increments of $0.01$. Ties are prohibited:
    do not output exactly $0.5$.
\end{itemize}

\paragraph{Output requirements.}
Output only one JSON object, without additional text or code fences.
The \texttt{basis} field must contain a concise explanation of no more
than 300 characters.
The following values and wording illustrate the format only:
\begin{verbatim}
{"p": 0.72, "basis": "Core grounds for the judgment."}
\end{verbatim}

\subsection{Task 2: Stage-Specific Score Prediction}

You are an experienced judge of competitive Chinese-language debate.
Using the competition's unified scoring standard, assign the specified
side an integer score from 1 to 10 for the specified stage.

\paragraph{Scoring standard.}
\begin{center}
\small
\begin{tabular}{|c|p{0.22\linewidth}|p{0.28\linewidth}|p{0.25\linewidth}|}
\hline
Score & Stage task completion & Choice of clash & Progress made \\
\hline
10 & Perfectly completed & Correct and important & Decisive progress \\
\hline
9 & Perfectly completed & Correct and important & Major progress \\
\hline
8 & Well completed & Correct and important & Substantial progress \\
\hline
7 & Well completed & Correct and important & Effective progress \\
\hline
6 & Well completed &
Correct and important / secondary &
Attempts at progress / effective progress \\
\hline
5 & Approximately completed & Correct but secondary &
Attempts at progress \\
\hline
4 & Approximately completed & Largely irrelevant &
Attempts at progress \\
\hline
3 & Not completed & Incorrect &
Attempts at progress \\
\hline
2 & Not completed & Incorrect & No attempts at progress \\
\hline
1 & Not completed & Incorrect & Counterproductive effect \\
\hline
\end{tabular}
\end{center}

\paragraph{Side-specific evaluation rules.}
\begin{itemize}
    \item \textbf{Monologue stages}
    (opening statements, substantive speeches, interim summaries,
    closing statements, or other speeches):
    assess how well the speaking side completed its task in this stage.
    \item \textbf{Questioning and cross-examination}:
    assess only the specified role.
    For the questioning side, evaluate the organization and progress
    of its attacks. For the responding side, evaluate its defense
    and handling of those attacks.
    \item \textbf{Direct exchanges}:
    evaluate the specified side's attack and defense.
    \item \textbf{Open debate---special rule}:
    evaluate the specified side's performance relative to its opponent.
    The larger the gap, the closer the stronger side's score should
    be to 10 and the lower the weaker side's score should be.
    \item The score must reflect performance within this stage only.
    Earlier stages provide argumentative context but must not
    contribute to the score.
\end{itemize}

\paragraph{Output requirements.}
Output only one JSON object, without additional text or code fences.
Use \texttt{p} for the integer score.
The \texttt{basis} field must contain a concise explanation of no more
than 300 characters.
The following values and wording illustrate the format only:
\begin{verbatim}
{"p": 7, "basis": "Core grounds for the stage score."}
\end{verbatim}

\subsection{Task 3: Best Debater Vote Allocation}

You are an experienced judge of competitive Chinese-language debate.
After reading the complete debate, allocate the votes for best debater.

\paragraph{Rules.}
\begin{itemize}
    \item Allocate exactly 9 votes in total, representing the combined
    votes of three judges with three votes each.
    \item Each debater must receive a nonnegative integer number of
    votes. The total must equal exactly 9.
    \item You may give all votes to one debater or distribute them
    among multiple debaters.
    \item Identify debaters exclusively by their speaking positions:
    \texttt{\{ROLES\}}.
    \item Evaluate each debater's actual contribution to this match:
    argument quality, performance in key exchanges, and influence
    on the course of the debate.
    \item Best debater votes need not follow the match result.
    Debaters on the losing side may also receive votes.
\end{itemize}

\paragraph{Output requirements.}
Output only one JSON object.
Keys must be speaking-position labels and values must be vote counts.
Positions receiving no votes may be omitted.
The following allocation illustrates the format only;
the total must equal 9:
\begin{verbatim}
{
  "Affirmative Speaker 3": 4,
  "Negative Speaker 2": 3,
  "Affirmative Speaker 1": 2
}
\end{verbatim}

Allocate the 9 votes now. Output JSON only.

\subsection{Task 4: Transcript Correction}

You are a text repair editor for debate transcripts.
The input consists of numbered utterance lines from one stage,
produced by Tencent Meeting automatic speech recognition (ASR).
It contains recognition errors and incorrect line breaks.

\paragraph{Tasks.}
\begin{itemize}
    \item \textbf{Correction}:
    correct only ASR recognition errors, including homophonic or
    near-homophonic substitutions, omitted characters, spurious
    characters, and obvious punctuation errors.
    \item \textbf{Resegmentation}:
    merge consecutive fragments from the same speaker into complete
    sentence units. A line containing multiple complete sentences
    may be split into multiple units.
    \item \textbf{No rewriting}:
    do not rephrase, add or remove content, supply words you merely
    assume are missing, or delete discourse particles.
    Preserve the conversational style.
    Leave uncertain characters unchanged.
\end{itemize}

\paragraph{Output format.}
\begin{verbatim}
{
  "units": [
    {
      "src": [1, 2],
      "text": "Corrected text."
    }
  ]
}
\end{verbatim}

\paragraph{Mandatory constraints.}
\begin{itemize}
    \item Every input line number, except those marked as context,
    must appear in the \texttt{src} field of exactly one unit.
    \item Each unit's \texttt{src} must contain consecutive line
    numbers belonging to the same speaker, as identified by the
    parenthesized \texttt{spk} label at the start of each line.
    \item Preserve the original order of units.
    \item Lines marked \texttt{[Context]} are provided only to aid
    understanding and must never appear in the output.
\end{itemize}

\subsection{Task 5: Judge's Verdict Compression}

Compress an exceptionally long judge's oral verdict into one paragraph
of no more than 300 Chinese characters.
Write in the judge's own voice, as a brief post-match explanation
of why the judge voted for the winning side and what decided the match.

\paragraph{Requirements.}
\begin{itemize}
    \item Base the content on the judge's verdict.
    Use the debate transcript only to clarify references.
    \item Write one continuous paragraph in the judge's first-person
    voice.
    \item Strictly observe the limit of 300 Chinese characters.
    \item Use the \texttt{Write} tool to write the result to:
\end{itemize}

\begin{verbatim}
${BASE}\outputs\\${mid}.json
\end{verbatim}

The file must contain a single JSON object with the following fields:
\begin{verbatim}
{
  "match_id": "${mid}",
  "n_judges_verdict": 2,
  "summary": "<compressed verdict>"
}
\end{verbatim}

Set \texttt{n\_judges\_verdict} to the actual number of judges whose
verdicts are provided: either 2 or 3.
The value above is illustrative.

\paragraph{Final response.}
Reply with exactly one line:
\begin{verbatim}
OK ${mid.slice(0, 12)}
\end{verbatim}

\subsection{Task 6: Chain-of-Thought Generation}

Read the following file:
\begin{verbatim}
${F}\rewrite_inputs\\${mid}.txt
\end{verbatim}

Rewrite its \texttt{[Judge's Verdict]} section into a
``chain of thought,'' strictly following the template and rules below.

\paragraph{Fixed structure.}
Fill in the bracketed placeholders for this match.
Where verbatim wording is required, reproduce all other wording
exactly as given.

\begin{enumerate}
    \item \textbf{Paragraph 1---copy verbatim, filling in only the
    motion and the two positions:}

    Okay, I now need to analyze the outcome of this debate and
    provide an outcome preference. The closer it is to 0, the more
    I favor the negative; the closer it is to 1, the more I favor
    the affirmative. The motion is: [motion]. The affirmative
    supports [affirmative position], and the negative supports
    [negative position].

    \item \textbf{Paragraph 2---copy verbatim:}

    First, I need to examine how both sides define the core
    concepts, then consider who substantiated their case in each
    clash, and only then determine the direction and degree of
    my preference.

    \item \textbf{Paragraph 3---definitions:}

    If the verdict discusses a definitional dispute, explain
    whose definition was more precise and who failed to dismantle
    the opponent's definition.

    If the verdict does not discuss a definitional dispute,
    state truthfully: ``There was little disagreement over
    definitions; this was not a central clash in the match.''
    Never invent a definitional dispute.

    \item \textbf{Paragraph 4---arguments:}

    Begin with: ``Then, consider the arguments.''

    Explain the affirmative's argumentative route and the
    negative's argumentative route, and identify whether they
    directly conflict or proceed in parallel.

    \item \textbf{Paragraph 5---clash declaration; use this exact
    sentence pattern:}

    Then come the clashes. I have identified what may be [N]
    clashes: the first is [clash name], and the second is
    [clash name].

    If there are three, append: ``, and the third is
    [clash name]'' before the final period.

    \item \textbf{Paragraph 6 onward---individual clashes:}

    Address each clash separately. Explain what is being
    contested, where the sides disagree, who substantiated
    their case, and who failed to answer.

    If the verdict explicitly treats a clash as tied or says
    the judge cannot decide, state faithfully:
    ``I judge this clash to be tied'' or
    ``I cannot determine the outcome of this clash.''
    Do not force a winner.

    You may use first-person judgments such as
    ``I need to assess this'' or
    ``They failed to answer this clash.''

    \item \textbf{Penultimate paragraph---overall assessment:}

    Begin with: ``Taking everything together, let me take stock.''

    Review how many points each side successfully established.
    Incorporate the meaning of the sentence under
    \texttt{[Calibration Guideline]} here, using your own words
    rather than copying it verbatim.

    \item \textbf{Final paragraph---use this exact closing pattern,
    filling in the placeholders:}

    Finally, [one or two sentences summarizing the direction
    and degree of the preference]. Let me think: I need to
    provide an outcome preference. The closer it is to 0,
    the more I favor the negative; the closer it is to 1,
    the more I favor the affirmative. Therefore, I will
    assign p a value of [p value].
\end{enumerate}

\paragraph{Mandatory constraints.}
\begin{itemize}
    \item All content must come from \texttt{[Judge's Verdict]}.
    Do not introduce arguments, examples, data, or names absent
    from the verdict.
    \item State the overall conclusion only in the final paragraph.
    Earlier paragraphs must not contain premature conclusions
    such as ``I vote for X.''
    \item Use plain prose paragraphs separated by one blank line.
    Do not use headings, numbered lists, or Markdown formatting
    in the generated text.
    \item Target 700--900 characters.
    If the source verdict is sparse, prefer a substantive
    500-character response to padded text.
    \item Use the $p$ value supplied in the input file,
    formatted to two decimal places.
\end{itemize}

\paragraph{File output.}
Use the \texttt{Write} tool to write a single JSON object to:
\begin{verbatim}
${F}\rewrite_outputs\\${mid}.json
\end{verbatim}

Use the following structure, placing \texttt{basis} before \texttt{p}:
\begin{verbatim}
{
  "match_id": "${mid}",
  "think": "<complete chain-of-thought text>",
  "basis": "<concise grounds for the judgment>",
  "p": 0.72
}
\end{verbatim}

The \texttt{basis} field must contain 80--100 characters and must
not begin with ``I vote for.''
Replace the illustrative numeric value with the supplied $p$ value.

\paragraph{Final response.}
Reply with exactly one line:
\begin{verbatim}
OK ${mid.slice(0, 10)}
\end{verbatim}

\bibliography{iclr2027_conference}

@inproceedings{zhao2023orchid,
  author    = {Zhao, Xiutian and Wang, Ke and Peng, Wei},
  title     = {{ORCHID}: A {Chinese} Debate Corpus for Target-Independent
               Stance Detection and Argumentative Dialogue Summarization},
  booktitle = {Proceedings of the 2023 Conference on Empirical Methods
               in Natural Language Processing},
  year      = {2023},
  pages     = {9358--9375},
  publisher = {Association for Computational Linguistics},
  address   = {Singapore},
  doi       = {10.18653/v1/2023.emnlp-main.582},
  url       = {https://aclanthology.org/2023.emnlp-main.582/}
}

@article{chen2026debate,
  author    = {Chen, Xunyu and Li, Yuanxia and Lee, One-Ki Daniel
               and Chin, Amita Goyal},
  title     = {Large Language Model-Empowered Debate Outcome Prediction
               via Argument Quality Assessment},
  journal   = {Group Decision and Negotiation},
  year      = {2026},
  volume    = {35},
  note      = {Article 31},
  doi       = {10.1007/s10726-026-09986-9},
  url       = {https://doi.org/10.1007/s10726-026-09986-9}
}

@inproceedings{yu2026defined,
  author    = {Yu, Tongzhou and Li, Mingjia and Qian, Hong
               and Wang, Wenkai and Zhang, Zongbao and Jiang, Yaoyu
               and Wang, Xiangfeng and Zhou, Aimin and Guo, Jiajun},
  title     = {{DEFINED}: A Data-Efficient Computational Framework
               for Fine-Grained Creativity Assessment in Debate Scenarios},
  booktitle = {Proceedings of the 32nd ACM SIGKDD Conference on
               Knowledge Discovery and Data Mining V.2},
  year      = {2026},
  pages     = {6292--6303},
  publisher = {Association for Computing Machinery},
  doi       = {10.1145/3770855.3817874},
  url       = {https://doi.org/10.1145/3770855.3817874}
}

@article{chen2026conch,
  author    = {Chen, Qianhe and Wang, Yong and Yu, Yixin
               and Zhu, Xiyuan and Yu, Xuerou and Wang, Ran},
  title     = {{Conch}: Competitive Debate Analysis via Visualizing
               Clash Points and Hierarchical Strategies},
  journal   = {IEEE Transactions on Visualization and Computer Graphics},
  year      = {2026},
  volume    = {32},
  number    = {1},
  pages     = {944--954},
  doi       = {10.1109/TVCG.2025.3634629},
  url       = {https://doi.org/10.1109/TVCG.2025.3634629}
}

@inproceedings{lan2026cedar,
  author    = {Lan, Tian and Li, Jiang and Yan, Rong and Bao, Feilong
               and Wang, Weihua and Gao, Guanglai and Su, Xiangdong},
  title     = {{CEDAR}: A {Chinese} Evaluation Dataset
               for Computational Argumentation},
  booktitle = {Proceedings of the 64th Annual Meeting of the Association
               for Computational Linguistics (Volume 1: Long Papers)},
  year      = {2026},
  pages     = {5247--5269},
  publisher = {Association for Computational Linguistics},
  address   = {San Diego, California, United States},
  doi       = {10.18653/v1/2026.acl-long.238},
  url       = {https://aclanthology.org/2026.acl-long.238/}
}

@inproceedings{debatrix2024,
  author    = {Liang, Jingcong and Ye, Rong and Han, Meng
               and Lai, Ruofei and Zhang, Xinyu and Huang, Xuanjing
               and Wei, Zhongyu},
  title     = {{Debatrix}: Multi-dimensional Debate Judge with
               Iterative Chronological Analysis Based on {LLM}},
  booktitle = {Findings of the Association for Computational Linguistics:
               ACL 2024},
  year      = {2024},
  pages     = {14575--14595},
  publisher = {Association for Computational Linguistics},
  address   = {Bangkok, Thailand},
  doi       = {10.18653/v1/2024.findings-acl.868},
  url       = {https://aclanthology.org/2024.findings-acl.868/}
}

@inproceedings{debatable2025,
  author    = {Sternlicht, Noy and Gera, Ariel and Bar-Haim, Roy
               and Hope, Tom and Slonim, Noam},
  title     = {Debatable Intelligence: Benchmarking {LLM} Judges
               via Debate Speech Evaluation},
  booktitle = {Proceedings of the 2025 Conference on Empirical Methods
               in Natural Language Processing},
  year      = {2025},
  pages     = {18850--18869},
  publisher = {Association for Computational Linguistics},
  address   = {Suzhou, China},
  doi       = {10.18653/v1/2025.emnlp-main.953},
  url       = {https://aclanthology.org/2025.emnlp-main.953/}
}

@article{eckstein2015bp,
  author  = {Eckstein, Justin and Bartanen, Michael},
  title   = {British Parliamentary Debate and the Twenty-First-Century Student},
  journal = {Communication Studies},
  year    = {2015},
  volume  = {66},
  number  = {4},
  pages   = {458--473},
  doi     = {10.1080/10510974.2015.1056916}
}

@article{schueler2025policy,
  author  = {Schueler, Beth E. and Larned, Katherine E.},
  title   = {Interscholastic Policy Debate Promotes Critical Thinking and College-Going: Evidence From Boston Public Schools},
  journal = {Educational Evaluation and Policy Analysis},
  year    = {2025},
  volume  = {47},
  number  = {1},
  pages   = {108--134},
  doi     = {10.3102/01623737231200234}
}

@article{zhong2002debating,
  author  = {Zhong, Yong},
  title   = {Debating with Muzzled Mouths: A Case Analysis of How Control Works in a Chinese Television Debate Used for Educating Youths},
  journal = {Media, Culture \& Society},
  year    = {2002},
  volume  = {24},
  number  = {1},
  pages   = {27--47},
  doi     = {10.1177/016344370202400102}
}

@misc{tiwari2025debatebench,
    title = {{DebateBench}: A Challenging Long Context Reasoning Benchmark For Large Language Models},
    author = {Tiwari, Utkarsh and
      Seth, Aryan and
      Mukherjee, Adi and
      Mer, Kaavya and
      Kavish and
      Kumar, Dhruv},
    year = {2025},
    eprint = {2502.06279},
    archivePrefix = {arXiv},
    primaryClass = {cs.CL},
    url = {https://arxiv.org/abs/2502.06279}
}
\bibliographystyle{iclr2027_conference}
\end{document}